\documentclass[runningheads]{llncs}
\usepackage{latexsym}
\usepackage{amssymb}
\usepackage[ruled, lined, linesnumbered, commentsnumbered, longend]{algorithm2e}
\usepackage{amsmath}

\usepackage{amsthm}
\usepackage{amsfonts}
\usepackage{amssymb}
\usepackage{comment}
\usepackage{graphicx}
\usepackage{subfig}

\usepackage{cleveref}
\DeclareMathOperator{\Tr}{Tr}

\newcommand*{\defeq}{\stackrel{\text{def}}{=}}
\newcommand{\set}[1]{\mathit{#1}}

\begin{document}
\title{GausSetExpander: A Simple Approach for Entity Set Expansion}
%
%
\author{A{\"i}ssatou Diallo\inst{1}, Johannes F{\"u}rnkranz\inst{2}}
\authorrunning{A{\"i}ssatou Diallo \and Johannes F{\"u}rnkranz}
%
\institute{Department of Computer Science, University College London, United Kingdom \and Computational Data Analytics, FAW,
Johannes Kepler University Linz, Austria \\
\email{a.diallo@ucl.ac.uk, juffi@faw.jku.at}}
\maketitle              
\begin{abstract}
Entity Set Expansion (ESE) is an important task in Natural Language Processing that aims at expanding a small set of entities into a larger one with items from a large pool of candidates. The problem implicitly requires the definition of the notion of similarity between the given entities and the candidates. In this paper, we propose \textit{GausSetExpander}, an unsupervised approach for the task of ESE based on optimal transport techniques. We propose to re-frame the problem as choosing the entity that best completes the input set. For this, we interpret a set as an elliptical distribution with a centroid which represents the mean and a dispersion that serves as the spread of the variance. The best candidate entity is the one that increases the spread of the set the least.
We analyze the strength and the weaknesses of the proposed solution in order to assess the validity of our proposed approach.

\keywords{Entity Set Expansion  \and Semantics \and Optimal Transport.}
\end{abstract}
\section{Introduction}
\label{sec:intro}

The Entity Set Expansion (ESE) task aims at expanding a small seed set e.g \{Paris, Berlin\} to a larger set of entities that belongs to the same semantic class, in this example $\textit{Capitals}$. Loosely speaking, the goal is to find all other entities in a given corpus that complete the original small set of  entities. The latter is called a \textit{seed set}. From a theoretical point of view, the ESE task can be seen as a problem of generalization from few examples. The steps that are necessary to successfully solve this problem are: (1) identifying the semantic class from the given seed set; (2) identifying similar examples from a large pool of items that fit the semantic class. This task is useful to several downstream applications such as question answering \cite{wang2008automatic}, taxonomy construction \cite{shen2018hiexpan}, relation extraction \cite{lang2013graph}, query suggestion \cite{cao2008context} or generation of user-guided dictionaries \cite{kohita2020interactive}. In order to perform well in the ESE task, the learner is required to have an approach of similarity between the elements that compose the seed set and the possible candidates for the expansion of the set.

In this paper, we propose an approach to tackle the ESE task based on pretrained embedding models and optimal transport techniques. It has been shown \cite{yaghoobzadeh2019probing} that pretrained embedding vectors contain semantic, syntactic and knowledge background that can be exploited for transfer learning tasks.  

One of the major challenges of ESE task is the limited (or lack thereof) of supervision. In fact, generally the set seed is too small for attempting fine-tuning and sometimes, the ground truth semantic class is an open set. 
From the literature, we can identify two main ways of solving the ESE task: a pattern-based approach  and a distributional approach. The first has as an objective mining revealing textual patterns in the corpus that signal the semantic class and extracting the correct entity from this patterns. On the other hand, the distributional approach relies on the assumption that similar words appear in similar contexts. These methods operate by representing each term in the vocabulary as an embedding vector that summarizes all the contexts the term appears in given a large corpus, and then look for terms
with vectors that are similar to those of the seed
entity. One of the main critiques to these approaches is that they consider
all occurrences of a term in the corpus when calculating its representation, including many contexts
that are irrelevant or non-informative to the concept which causes noise in the corpus.  

We propose an algorithm that is more similar to the distributional approach, although our algorithm has some notable differences. \emph{GausSetExpander}, proceeds in an iterative manner across all terms in a vocabulary $V$, extracted from a corpus $D$. Given an initial seed set $S_0$, and two terms embedded with a pretrained embedding model, we aim at finding the term that completes better the original set seed $S_0$. For doing this, we simply produce two different expanded sets from the simple concatenation of the previously two random terms to the seed set and we evaluate the better expansion. 
The evaluation of the new obtained sets relies on techniques related to optimal transport, elliptical embeddings and set clustering, which are concepts this paper is treating. 
In a nutshell, we leverage the fact that one of the most common statistics of a set $S$ is its centroid $c(S) \in \mathbb{R}^d$, represented as:
\begin{equation}
    c(S) = \frac{1}{|S|}\sum_{i \in S} x_i
\label{eq:centroid_setext}
\end{equation}
This implies that, it is possible to approximate a set of vectors to its centroid. Another key element of our approach comes from the literature of Gaussian embeddings, which has demonstrated to provide a better and richer representation for items in the latent space. In fact, a Gaussian embedding is characterized by two parameters, the mean vector which represents the location in the latent space and the covariance matrix, which encodes the dispersion around the location vector, that is to say the uncertainty of the representation. 

We make the hypothesis that given an original seed set and two candidate items to expand it, the better entity is the one that causes the less increase in the dispersion of the set. This hypothesis is the key element of GausSetExpander. 
Additionally, ESE is a challenging task for the lack of available labels. 
For this reason, the pretrained embedding is a key element, because at each iteration we produce a weak label based on the cosine similarity between the centroid of the seed set and each candidate term. Finally, we rank the scores in order to identify candidates term from the vocabulary to generated the expanded set. 
To summarize, in this study we propose an iterative approach based on Gaussian representation for the ESE task to expand a seed set. We conduct experiments to verify our hypotheses and show the effectiveness of GausSetExpander. 

The paper is organized as follows: we first present some related works that tackles the ESE task, pre-trained embedding and learning from sets. Then, we present the problem formulation and introduce the necessary notation used throughout the paper. Next, we introduce our approach \textit{GausSetExpander} as well as a natural baseline derived from our hypothesis. Finally, we present our experimental setup and discuss the results obtained.

\section{Related Work}
\subsection{Entity Set Expansion} The Entity Set Expansion is a task that can be seen as a intermediate step for many problems and applications such as question answering or literature search. It can be considered a weakly supervised task in which a small set of entities, the seed set is used as supervision to retrieve the other entities in order to expand the small set. A first categorization can be made based on the corpus used for the ESE task. In fact, the used corpus can be either limited such as in \cite{shi2014probabilistic,shen2017setexpan}, or an open corpus relying on a search engine for the web \cite{wang2007language}.
It was common for entity set expansion systems such as Google Sets \cite{tong2008system} and SEAL \cite{wang2008automatic} to submit a query made of seed
entities to a general-domain search engine and then extract new entities from retrieved web pages. 
Web-based methods for
set expansion \cite{chen2016long} extract entities from documents retrieved by
a search engine. The documents are chosen with respect to the query obtained from seed entities. However, this type of
approaches demand an external search engine adapted to seed-oriented data, which could be
costly. In fact, such methods impose significant run-time overhead and they make the assumption that top-ranked web pages contain other entities of the set,
which might not necessarily be exact. 

As a solution, more recent works propose to
expand the seed set processing a corpus offline without the need to prompt an online search engine.
These corpus-based set expansion methods can be
categorized into two general approaches: (1) entity ranking which calculates entity similarities and ranks all entities \cite{mamou2018term,pantel2009web,kushilevitz2020two}, and (2) iterative bootstrapping
whose goal is to bootstrap the seed entity set by iteratively selecting context features and ranking new
entities \cite{shen2017setexpan,huang2020guiding,rong2016egoset} with respect to the original seed set. The approach proposed in this paper belongs to the first category. 

\subsection{Pre-training} Early works in computer vision on ImageNet \cite{russakovsky2015imagenet} have shown the advantages of pretrained models \cite{huang2017speed,he2017mask}. In Natural Language Processing, pretrained embeddings such as Glove \cite{Pennington2014GloVe:Representation} or Word2Vec \cite{Mikolov2013DistributedP} have also proven to be highly effective in several tasks. In addition to the embeddings, the pretrained language models are currently vastly used such as BERT \cite{Devlin2019BERT:Understanding}.

\subsection{Set encoding}
Machine learning on sets can be divided into different subgroups depending on the nature of the input and output (e.g vector-to-set, set-to-set, set-to-sequence). Our work falls under the category of vector-to-set such as \cite{Zhang2020FSPool:,zhang2019dspn}. 
 Notable works in vector-to-set literature are suited to tasks such as object detection, taking as input a feature representation of an images and producing set of coordinates for the bounding boxes. Loosely related are also methods like \cite{zhang2019permoptim} that learn a permutation matrix for sets of items. Once the permutation matrix is learned, it is applied to the input set in order to turn the set into an (ordered) sequence. Once again, our method differs because the output set is not ordered, hence not a sequence. 
 
\subsection{Probabilistic Embeddings} 
The work of \cite{vilnis2014word} established a new trend  in the representation learning field
by proposing
to embed data points, in this case words, as probability distributions in $\mathbb{R}^d$.  Representing objects in the latent space as probability distributions allows more flexibility in the representation and even express multi-modality. 
In fact, point embeddings can be considered as a special case of probabilistic embeddings, namely a Dirac distribution, where the uncertainty is collapsed in a single point.
In the above-mentioned work \cite{vilnis2014word}, the metric used is KL divergence. However, this metric has a drawback: when variances of the probabilistic embedding collapse, the measure does not coincide with the Euclidean metric between point embeddings. Loosely speaking, the KL divergence and the $\ell_2$ distance between two probability measures diverge to infinity when the variances become too small. In addition, the KL divergence does not behave well when the two compared distributions have little or no overlap. It has been shown by \cite{muzellec2018generalizing} that the Wasserstein metric is a better metric to compare probabilistic embeddings.

\section{Problem Formulation}
\subsection{Terminology} In this section, we formally introduce the concepts necessary for this study as well the Entity Set Expansion problem. 
An \textbf{entity} is a string, that can be either a word or a phrase, that corresponds to a real-world entity in designated category. This category is better defined as a \textbf{semantic class} which is a set of entities that share a common characteristic, for example \textit{fruits} or \textit{animals}. An \textbf{entity set}, is a set of terms that refers to the same real-world entity. More specifically, \textit{UK} and \textit{United Kingdom} refers to the same entity, hence are an \textit{entity synset}. It is worth mentioning that an entity synset can be a singleton. Finally, a \textbf{vocabulary} is defined as the list of terms that is derived from a given corpus. 

\subsection{Problem Formulation} Given a text corpus $D$, a vocabulary $V$ extracted from $D$, a seed set of size $n$ (usually small) of terms $S_0=[e_1, e_2, ... e_n]$ from the same semantic class $C$, the Entity Set Expansion task aims at expanding $S_0$ with terms from the vocabulary $V$ from the same semantic class $C$. In practice, the goal is to rank all available entities and said ranking must have the terms that best expand $S_0$ at the top of the ranking.

\section{Approach}
In this section, we first lay down the notation as well as the mathematical background used for our hypothesis. Then we describe our hypothesis and the algorithm we proposed: \emph{GausSetExpander}.

\subsection{Notation}

$\mathcal{S}^{d}_+$ is the set of all positive definite matrices. 
In the scope of this work, we only focus on Gaussian distributions which belong to the family of parameterized probability distributions $z_{h,\mathbf{a}, \mathbf{A}}$ having a location vector $\mathbf{a} \in \mathbb{R}^d$ which represents the shift of the distribution, a scale parameter $\mathbf{A} \in \mathcal{S}^{d}_+ $, that represents the statistical dispersion of the distribution, and a characteristic generator function $h$. This is the same hypothesis that have been used throughout the paper. Specifically, for Gaussian distributions, the scale parameter coincides with the covariance matrix $var(z_{h,\mathbf{a}, \mathbf{A}}) = \mathbf{A}$. From now on, we denote Gaussian 
distributions (or embeddings) as $z_{(h,\mathbf{a}, \mathbf{A})} = \mathcal{N}(\mathbf{a}, \mathbf{A})$. 

Finally, when stated $\Tr$, we indicate the trace operator, that is the sum of the diagonal elements of a matrix.

\subsection{Mathematical Background}
In this section, we explicitly describe few mathematical concepts that are useful for understanding our approach.

\subsection{Set representation}
First, as briefly mentioned in \Cref{sec:intro}, one key element in our approach is the notion of centroid of a set from eq. \eqref{eq:centroid_setext}. In fact an (aggregation) function that takes as input a set needs to have the property of permutation-invariance. That is to say that this function should not rely on the arbitrary order of the elements of the input set. 
Multiple operations abide by this rule, for example, sum, average, min or max. This allows us to obtain a representation of the set and its elements regardless of the order in which of the set elements are presented. We made the choice of using the \emph{centroid}, the center of mass described by eq. \eqref{eq:centroid_setext}.

\subsection{Distributional Embedding}

\begin{figure*}[h]
\centering
\subfloat[A Gaussian set embedding with a large spread among the items.]{\includegraphics[scale=0.2]{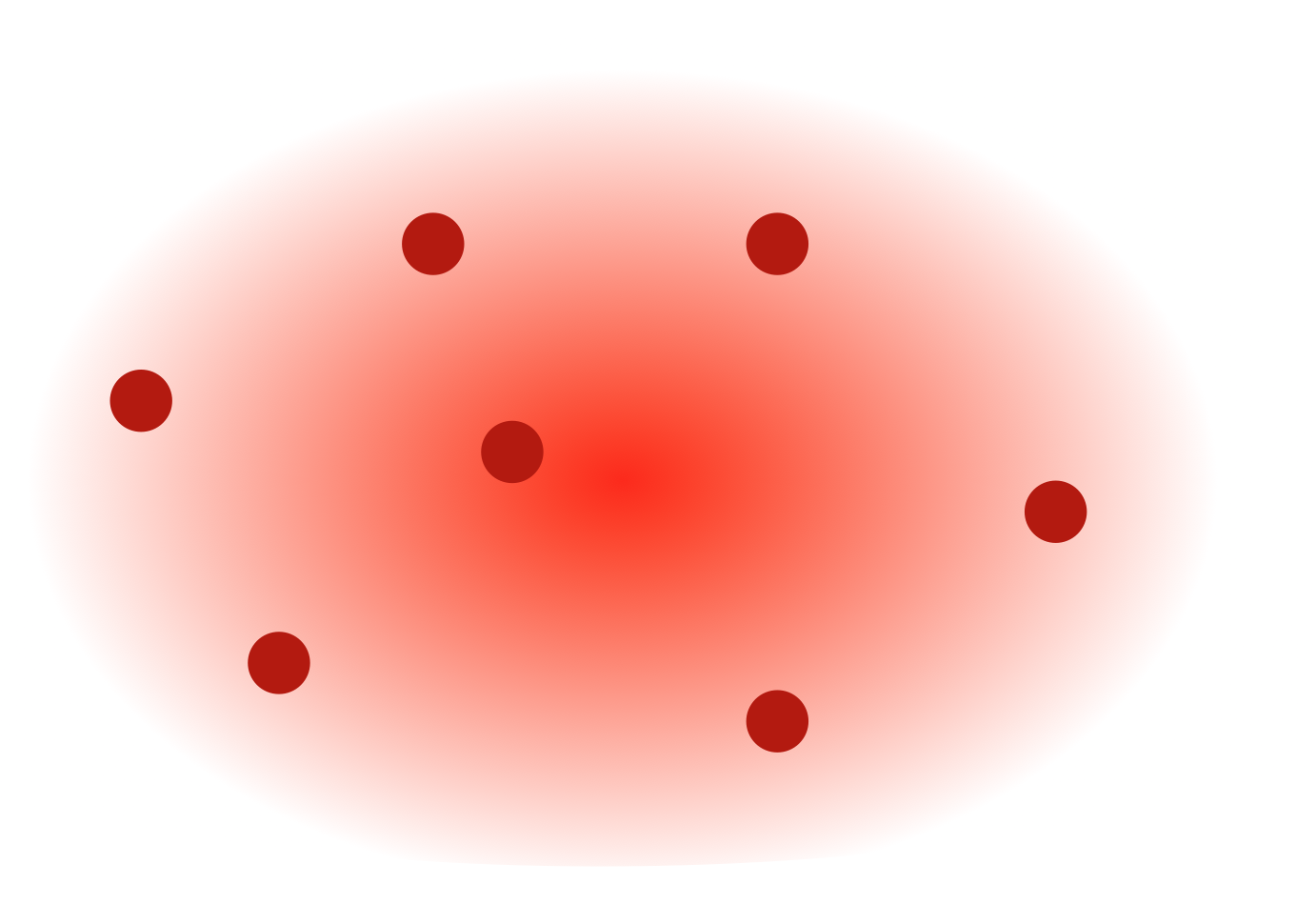}}
\hspace{3cm}
\subfloat[A more concentrated variance indicates a higher degree of homogeneity of the items within the set.]{ \includegraphics[scale=0.25]{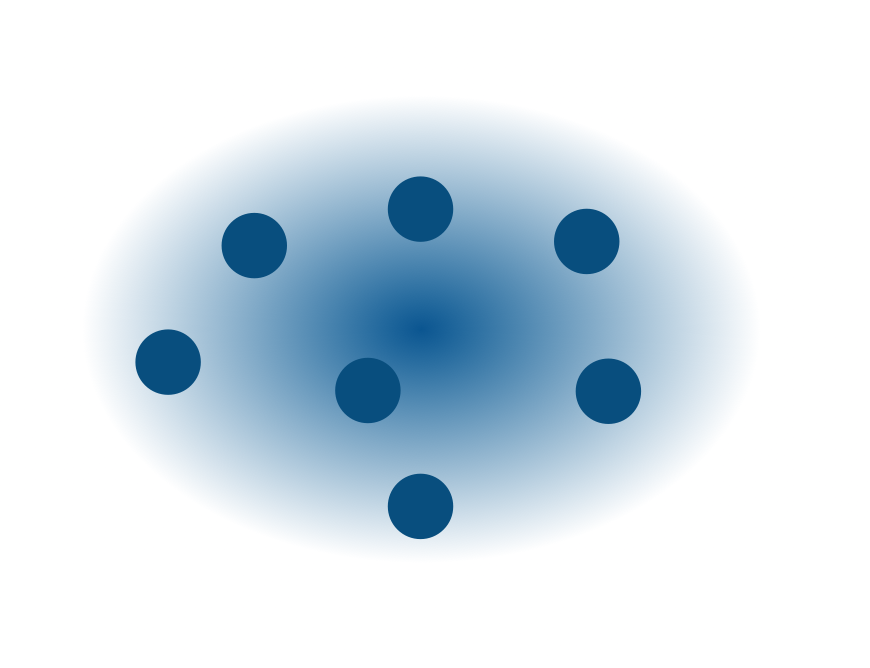}}
\caption{Schematization of the main idea of the distributional embeddings for sets. The Gaussian set embedding is represented by the centroid of the vector representation of the single items and the spread of the items within the set.}
\label{fig:distr_gaussian}
\end{figure*}

Then, we need to go from the point-vector representation of an item to a distributional representation in the embedding space. We choose to focus on Gaussian distributions, particularly diagonal Gaussian distributions. 
To that end, we chose the Gaussian distribution $\mathcal{N}(\mu, \Sigma)$ characterized by a location vector $\mu$ and a covariance matrix $\Sigma$, i.e., we associate each set $\set{X}$ with a distribution $\mathbf{z}_{\set{X}} = (\mu_\set{X},\Sigma_\set{X})$, where 
$\mu_\set{X}= \bar{\mathbf{y}}$,
i.e., the centroid~\eqref{eq:centroid_setext} of the set, and a diagonal covariance matrix $\Sigma_{\set{X}}$.
Overall, such a distributional embedding for sets 
allows to obtain a meaningful representation for each set, while being bounded by the components of the sets themselves. 
For doing this we propose to use a set encoder, that takes as input a vector, mainly the centroid of the set and outputs a location vector $\mu$ and the diagonal of a covariance matrix $d_{\Sigma}$ as shown in \cref{fig:model_gaussetext}. An illustration of these concepts can be seen in \cref{fig:distr_gaussian}

\subsection{The 2-Wasserstein distance}

Finally, we need to describe the scoring function that takes as input two Gaussian representation of sets and assess which one is the best representation. As stated earlier, the best set is the one that have the smallest dispersion. We propose to take advantage of the $\Tr$, trace operator for the scoring function of the sets. This is derived from the optimal transport theory, in particular the Wasserstein distance.

In Optimal Transport (OT) theory, the Wasserstein or Kantorovich–Rubinstein metric is a distance function defined between probability distributions (measures) on a given metric space $M$. The squared Wasserstein metric for two arbitrary probability measures $\mu, \nu \in \mathcal{P}(\mathbb{R}^d)$ is defined as: 

\begin{equation*}
    W_2^2(\mu, \nu) \defeq{} \inf_{X\sim \mu, Y\sim \nu } \mathbb{E}_{\|X-Y\|^2}
\end{equation*}
In the general case, it is difficult to find analytical solutions for the Wasserstein distance. However, a closed form solution exists in the case of Gaussian distributions. Let $\alpha \defeq{}\mathcal{N}(\mathbf{a}, \mathbf{A})$ and $\beta \defeq{}\mathcal{N}(\mathbf{b}, \mathbf{B})$, where $\mathbf{a},\mathbf{b} \in \mathbb{R}^d$ and $\mathbf{A}, \mathbf{B} \in \mathcal{S}^d_+$ are positive semi-definite. Hence:
\begin{equation}
\label{eq:wasserstein}
    W^2_2(\alpha, \beta) = \|\mathbf{a}-\mathbf{b}\|^2 + \mathfrak{B}^2(\mathbf{A}, \mathbf{B}) 
\end{equation}
where $\mathfrak{B}^2$ is the \textit{squared Bures metric} \cite{dittmann1999explicit}, defined as:
\begin{equation}
    \mathfrak{B}^2(\mathbf{A}, \mathbf{B}) \defeq{} \text{Tr}(\mathbf{A} + \mathbf{B} -2(\mathbf{A}^{\frac{1}{2}}\mathbf{B}\mathbf{A}^{\frac{1}{2}})^{\frac{1}{2}})
\end{equation}
When $\mathbf{A}=\text{diag }\mathbf{d_A}$ and $\mathbf{B}=\text{diag }\mathbf{d_B}$ are diagonal, $W^2_2$ simplifies to the sum of two terms:
\begin{equation}
     W^2_2(\alpha, \beta) = \|\mathbf{a}-\mathbf{b}\|^2 + \mathfrak{h}^2(\mathbf{d_A}, \mathbf{d_B})
\label{eq:wasserstein_ese}
\end{equation}
where $\mathfrak{h}^2(\mathbf{d_A},\mathbf{d_B}) \defeq{} \|\sqrt{\mathbf{d_A}} - \sqrt{\mathbf{d_B}}\|^2$ is the \textit{squared Hellinger distance} \cite{beran1977minimum} between the diagonal $\mathbf{d_A}$ and $\mathbf{d_B}$.

\section{GausSetExpander}

In the following section, we present the proposed approach for the Entity Set Expansion task. 
We propose an iterative approach that takes as input a small seed set and two candidates term and outputs two scores that indicates the best alternative. In a nutshell, our main hypothesis is that given two candidate entities for expanding the seed set, the best hypothesis will "fit" better the input set. We choose to represent this fit by considering the dispersion that the addition of a new term will cause to an existent set. A well suited mathematical notion for representing the dispersion around a center is the covariance matrix of a Gaussian distribution. That is the first key element of GausSetExpander. \Cref{algo:gausset_algo} describes our approach. 

\begin{figure}[h]
\centering
\includegraphics[scale=0.4]{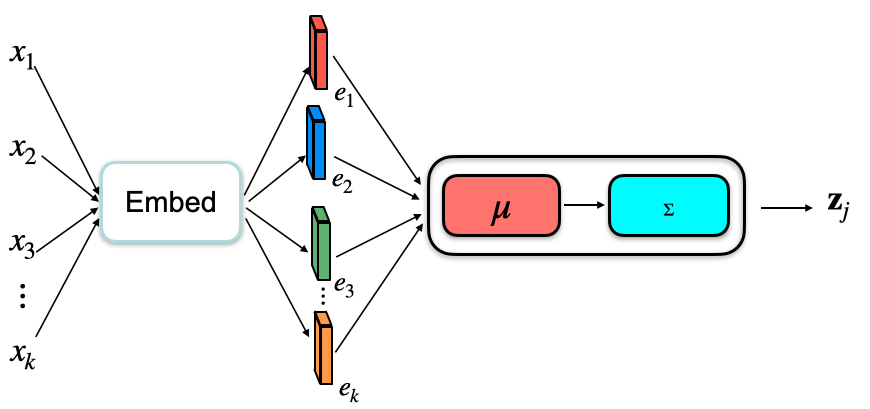}
\caption{Overview of the set encoder that takes as input a set of terms and outputs a tuple of mean vector and a covariance matrix for the Gaussian distribution.}
\label{fig:model_gaussetext}
\end{figure}

\subsection{Set Encoder}
Given a seed set, the first step is to encode it as a tuple of location vector and covariance matrix in order to have the the dispersion before adding the candidate terms for the expansion. For this, 
we represent the element-wise embedding function $\phi(.)$ as a deep neural network, as illustrated in Figure~\cref{fig:model_gaussetext}. 
This network takes as input a set. We again choose to only focus on diagonal covariance matrices for the sake of simplicity. 
A first deep encoder, namely a 2-layer MLP with Relu, $\phi_\theta(\cdot)$ maps these random inputs into $d$-dimensional outputs. These are then aggregated to $\mu_\theta(\cdot)$ and fed to produce the variance $\Sigma_\theta(\cdot)$, a function which is again represented with a deep forward network. 

\subsection{Scoring function}
After describing the step to obtain the Gaussian representation of the sets, we illustrate the scoring process. In practice, most of the proposed methods for solving the ESE problem return a (top-k) ranking of the vocabulary rather than a fixed set. Then the evaluation is done on the returned ranking. Ideally, all terms that belongs to the semantic class identified by the seed set should be ranked higher. 

From the Wasserstein distance in eq. \eqref{eq:wasserstein_ese}, we derive the scoring function for two given expanded sets encoded as Gaussian distributions is:
\begin{equation}
score(S_i, S_j) = W((\mu_i, \Sigma_i), (\mu_j, \Sigma_j))
    \label{eq:score}
\end{equation}
This scoring function can be interpreted as point-wise mutual information between the candidate item and the set to be expanded. 

We stated earlier that GausSetExpander proceeds in stages. It first encodes the seed set $S_0$ as a tuple of vectors $\mathbf{\mu}_0$ and $\mathbf{\sigma}_0$. 
Then, it appends each candidate to the seed set $S_0$ to obtain two new sets $S'$ and $S''$, which are transformed in Gaussian distributions as well. 
It is worth mentioning, that the weights of the set encoder shared among candidates. Finally, the scoring function proceeds as follows: 
\begin{gather}
    score(i|S_0) = W((\mu_0, \Sigma_0), ((\mu', \Sigma'))) \\
    score(j|S_0) = W((\mu_0, \Sigma_0), ((\mu'', \Sigma'')))
\end{gather}
where $(\mu_0, \Sigma_0)$ are the parameters of the Gaussian encoding of $S_0$. $(\mu', \Sigma')$ and $(\mu'', \Sigma'')$ are the parameters for $[S_0, i]$ and $[S_0, j]$. $[]$ indicates the concatenation.

\begin{figure*}[h]
\centering
\subfloat[The expanded set $S'$ has a small increase in the dispersion after adding the entity $i$ to the seed set $s_0$. ]{\includegraphics[scale=0.13]{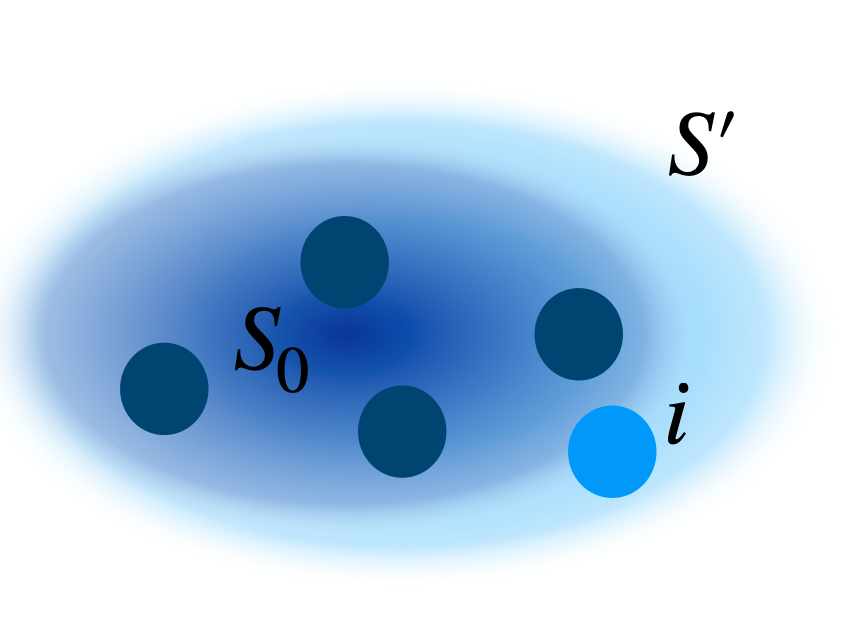}\label{fig:gaus_ext_pos}}
\hspace{3cm}%
\subfloat[The expanded set $S''$ has a greater increase in the dispersion after adding the entity $j$ to the seed set $s_0$.]{\includegraphics[scale=0.13]{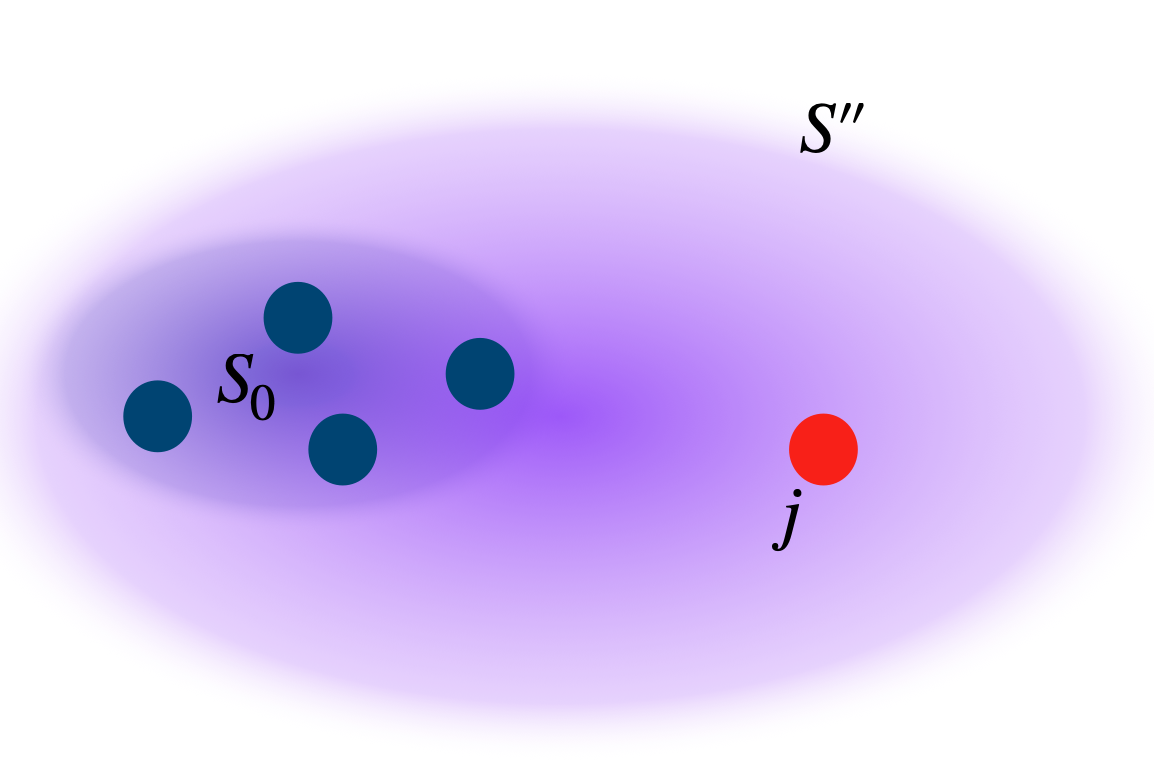}\label{fig:gaus_ext_neg}}
\caption{Illustration of the hypothesis under GausSetExpander. Given two candidate entities for the seed set $S_0$, the best option is the one that will induce the least increase in the dispersion of the generated set. }
\label{fig:mesh1}
\end{figure*}

\subsection{Loss function}
For learning the score, we utilize the large-margin classification loss:
\begin{equation}
    l(i,j|S_0) = \max(0, s(i|S_0) - s(j|S_0) + \Delta(i,j) )
\end{equation}
Loosely speaking, this loss function is to ensure that $s(i|S_0)) > s(j|S_0) + \Delta(i,j)$  whenever $i$ should be preferred for expanding $S_0$ over $j$.

\subsection{Weak supervision} Per definition, the ESE task is challenging for the lack of proper supervision under the form of ground truth labels. Several works in the area rely on pre-trained language models or pre-trained embedding models to deliver the semantic, syntactic and background knowledge to provide weak labels. We proceed in the same manner, and use Glove \cite{Pennington2014GloVe:Representation} for encoding the terms extracted from the corpus. Moreover, we assume that the seed terms are part of vocabulary. In order to generate, weak labels for training the scoring function we leverage the distributional hypothesis of the language models. In fact, that hypothesis states that similar words appear in similar contexts. For this reason, given two candidates terms $i$ and $j$, we extract their context as well, $c_i,$ and $c_j$ respectively. Then we use the simple cosine similarity to induce a weak label $l$:
\begin{equation}
l(S_0, c_i, c_j) =  \left\{
    \begin{array}{ll}
        +1 & \mbox{if } R(c_i|S_0) > R(c_j|S_0) \\
        -1 & \mbox{if } R(c_i|S_0) < R(c_j|S_0)
    \end{array}
\right.
\label{eq:label}
\end{equation}
The function $R$ is defined as follows:
\begin{equation}
R(x| S_0) = \max_{s_i \in S_0} \mathit{cosine\_sim}(x, s_i)
\end{equation}

\begin{algorithm}[h]

    \SetKwInOut{KwIn}{Input}
    \SetKwInOut{KwOut}{Output}

    \KwIn{corpus $D$, vocabulary $V$, set $S_0$}

    \For{$t \leftarrow 0$ \KwTo $|D|$}{
    Get i, j, $c_i$, $c_j$ from $D$ \\
    Embed i and j to $\mathbf{e_i} \in \mathbb{R}^d$,  $\mathbf{e_j} \in \mathbb{R}^d$ \\
    Embed terms in $S_0$ to $[\mathbf{e_0}, ..., \mathbf{e_n}]$ \\
    Encode $S_0 \leftarrow \mathcal{N}(\mu_0, \Sigma_0)$: \newline
    $c(S_0) = \frac{1}{n} \sum_i e_i$
    $\mu_0$= $\phi_\mu(c(S_0))$ \newline
    $\Sigma_0$ = $e^{\frac{1}{2} \phi_\Sigma(c(S_0))}$ \\
    Append $\mathbf{i}$ and $\mathbf{j}$ to $S_0$: \newline
    $S_t^{'} = [S_0, e_i]$ and $S_t^{''} = [S_0, e_j]$ \\
    Compute cosine similarity: \newline
    $sim_1 = \cos(c(S_0), c(c_i))$ and $sim_2 =\cos(c(S_0), c(c_j))$  \\
    \eIf{$sim_1$ > $sim_2$}{$l = 1$}{$l = -1$}
    Encode $S_t^{'}$ and $S_t^{''}$: \newline 
    $S_t^{'} \leftarrow \mathcal{N}(\mu_t^{'}, \Sigma_t^{'})$ \newline 
    $S_t^{''} \leftarrow \mathcal{N}(\mu_t^{''}, \Sigma_t^{''})$ \\
    Compute the score: \newline
    $score(i|S_0) = W((\mu_0, \Sigma_0), (\mu_i, \Sigma_i))$ \newline
    $score(j|S_0) = W((\mu_0, \Sigma_0), (\mu_j, \Sigma_j))$ \newline
    $l(i,j|S_0) = \max(0, s(i|S_0) - s(j|S_0) )$
    }
    \caption{GausSetExpander}
    \label{algo:gausset_algo}
\end{algorithm}

\begin{algorithm}[h]
    \SetKwInOut{KwIn}{Input}
    \SetKwInOut{KwOut}{Output}

    \KwIn{corpus $D$, vocabulary $V$, set $S_0$}

    \For{$t \leftarrow 0$ \KwTo $|D|$}{
    Get i, j, $c_i$, $c_j$ from $D$ \\
    Embed i and j to $\mathbf{e_i} \in \mathbb{R}^d$,  $\mathbf{e_j} \in \mathbb{R}^d$ \\
    Embed terms in $S_0$ to $[\mathbf{e_0}, ..., \mathbf{e_n}]$ \\
    Encode $S_0$ as its centroid: \newline
    $c(S_0) = \frac{1}{n} \sum_i e_i$
    Append $\mathbf{i}$ and $\mathbf{j}$ to $S_0$: \newline
    $S_t^{'} = [S_0, e_i]$ and $S_t^{''} = [S_0, e_j]$ \\
    Compute cosine similarity: \newline
    $sim_1 = \cos(c(S_0), c(c_i))$ and $sim_2 =\cos(c(S_0), c(c_j))$  \\
    \eIf{$sim_1$ > $sim_2$}{$l = 1$}{$l = -1$}
    Encode $S_t^{'}$ and $S_t^{''}$ as their centroids: \newline 
    $S_t^{'} \leftarrow c(S_t^{'})$ \newline 
    $S_t^{'} \leftarrow c(S_t^{''})$ \\
    Compute the score: \newline
    $score(i|S_0) = l^2(c(S_0) - c(S_t^{'}))$ \newline
    $score(j|S_0) = l^2(c(S_0) - c(S_t^{'}))$ \newline
    $l(i,j|S_0) = \max(0, s(i|S_0) - s(j|S_0) )$
    }
    \caption{CentroidSetExpander}
    \label{algo:centroset_algo}
    
\end{algorithm}

\section{Experiments}
In this section we describe our experimental setting as well as the main results obtained.

\subsection{Dataset} We used English Wikipedia as the corpus. Following, \cite{kushilevitz2020two}, we use for evaluation 7 datasets of closed sets. These sets are the National football league teams (NFL,
size:32), Major league baseball teams (MLB, size:30), US states (US, size:50), Countries (Cntrs, size:195),
European countries (Euro, size:44) Capital cities (Caps,
size:195) and Presidents of the USA (Pres, size:44) and one open class set: Music Genres
(Genre). 

\begin{table*}[t]
\centering
\begin{tabular}{ |l|ccccccc|c|c|} 
 \hline
 Method & NFL & MLB & Pres & US & Cntrs & Euro & Caps & Music & Avg \\
  \hline
 \textbf{SetExpander} & .54 & .45 & .33 & .55 & .55 & .61 & .14 & \textbf{.99} & .52 \\
 \textbf{CB} & \textbf{.98} & \textbf{.97} & \textbf{.70} & .93 & .74 & .46 & .21 & .67 & .71 \\
 \textbf{Cent-SetExp.} & .89 & .67 & .10 & .70 & .64 & .46 & .19 & .62 & .54 \\
\textbf{GausSetExp.} & \textbf{.98} & .69 & .14 & \textbf{.98} & \textbf{.76} & \textbf{.74} & \textbf{.34} & .98 & .70 \\

 \hline
\end{tabular}
\vspace{0.5cm}
\caption{Set expansion results. We report the MAP@k for different k. Results are averaged over 3 random seed of cardinality 3. }
\label{tab:results}
\end{table*}

\subsection{Setup and Evaluation} We rely on 300-dimensional Glove for obtaining the pre-trained embedding. Moreover, we fix the maximum number of terms to 200,000 most frequent terms. Our approach has also a learning module, for which we need to define the following hyperparameters $l_r=1e-3$ as the learning rate, batch size $512$, $h=64$ as the hidden size. Following other works in the literature, we chose the Mean Average Precision (MAP) at different positions k for evaluating our approach:
\begin{equation}
    MAP@k = \frac{1}{|V|}\sum_{t \in V} AP(S_{exp}, S_{gt})
\end{equation}

where $AP(S_{exp}, S_{gt})$ is the average precision at position k given the ranked expanded set $S_{exp}$ and the ground-truth set $S_{gt}$.
The training and the evaluation is done with random set seeds of cardinality 3 and the results presented are averaged over random seeds. We follow the recommendation of the authors of the datasets used for evaluation and fixed the size of the expanded set to 200 for smaller sets and to 350 for bigger sets that have size $>$ 100. The open set of music genre is evaluate with $\text{MAP}_{70}$.

\subsection{Baseline methods}
We compare our approach to the following corpus based ESE approaches: \textbf{SetExpander} \cite{mamou2018term}, a method based on entity ranking based on multi-context entity similarity defined on multiple embeddings, \textbf{Category Builder} \cite{mamou2018term}. Both this approaches are distributional, which is the category to which our method belongs to. For the sake of gaining a better insight of our hypothesis, we add a natural comparison approach which consists in our same architecture but the encoding of the sets remain the centroid representation. This comparison will act as an ablation experiment which we perform to assess the benefit or lack thereof of the Gaussian representation of the sets. We call this last baseline, Centroid-SetExpander and we described it in \cref{algo:centroset_algo}.

\subsection{Results} We assess the set expansion performance by comparing our approach to the baseline methods. \Cref{tab:results} presents the quantitative evaluation of our model. We can notice that GausSetExpander achieves comparable performance with current state-of-the-art methods. This demonstrates the validity of our hypothesis. When doing a fine grained analysis, we notice that GausSetExpander performs particularly poorly in the semantic class corresponding to the "Presidents". When investigating the results retrieved, we hypothesize that our algorithm probably fails in recognising the acronyms. In fact, a typical synset for the class "Presidents" will be \textit{"John\_Fitzgerald\_Kennedy, John\_F.\_Kennedy, John\_F\_Kennedy, John\_Kennedy, Kennedy, JFK"}. We hypothesize that probably a better encoding of the entities and tokens could resolve this problem. 

An emblematic example could be that, among the top retrieved entity for the class "Presidents" we can find "Hillary" or "Barack" or even Presidents of other nations which were not present among the ground-truths. This let us think that the results might not be as bad as the numerical evaluation shows. \Cref{tab:case_study} showcases few examples of top ranked entities but rejected. 

One reason for this phenomenon could be the fact that the weak label is generated from a very noisy procedure. In fact, it is the result of the cosine similarity between the centroid of the set seed and the contexts of the two candidates entities. We hypothesize that a better method for generating a weak label than the naive cosine similarity could be beneficial. This leads to a a second point that it is worth mentioning which is that, our algorithm is very sensible to the amount of data. In fact, after a threshold which we empirically estimated around 50,000 articles, with a vocabulary of around 500,000 terms, we notice that the labels become extremely noisy and the overall performance degrades rapidly. 

Finally, we compare \textit{GausSetExpander} to \textit{CentroidSetExpander}. We notice that the performance of the latter is worse in all datasets, and it declines even further for the semantic class "Presidents". This confirms that our initial hypothesis of considering the Wasserstein Distance between the sets is effective for the task of Entity Set Expansion.

\begin{table*}[t]
\centering
\begin{tabular}{ |l|c|} 
 \hline
 \textbf{Seed set} & \textbf{Top-k results} \\
  \hline
 \textit{Presidents} & ['mccain', 'barack', 'hillary', 'gore', 'fahmy', 'letterman'] \\
 \textit{Capitals} & ['georgia', 'ababa', 'petersburg', 'berliner', 'sapienza'] \\
 \textit{Countries} & ['maldive', 'federation', 'duchy', 'bissau', 'herzegovina'] \\
  \textit{MLB} & ['milwaukee', 'hammonds', 'packers', 'tigers', 'cardinals'] \\
\textit{Euro} & [''herzegovina', 'yugoslavia', 'slovak', 'tirana', 'uzbekistan'] \\
 \hline
\end{tabular}
\vspace{0.5cm}
\caption{Case study of entities retrieved but rejected for the least performing semantic classes.}
\label{tab:case_study}
\end{table*}

\section{Conclusion}
We introduce an iterative and distributional approach for the task of Entity Set Expansion. Our method is based on encoding sets of vectors as Gaussian distributions. These probabilistic representations are learned from the centroid of the sets and are useful because they allow to represents the sets as a tuple of location and dispersion. In fact, our main hypothesis states that given to candidates entities for a set, the best candidate will be the entity that will cause the least increase in the dispersion, represented as the covariance matrix. Finally, we use a scoring function based on the Wasserstein distance. The quantitative evaluation on benchmark datasets demonstrate the effectiveness of our approach.

\bibliographystyle{splncs04}
\bibliography{custom,word_embedding}
\end{document}